\documentclass[lettersize,journal]{IEEEtran}
\usepackage{amsmath,amsfonts}
\usepackage{algorithmic}
\usepackage{array}
\usepackage[caption=false,font=normalsize,labelfont=sf,textfont=sf]{subfig}
\usepackage{textcomp}
\usepackage{stfloats}
\usepackage{url}
\usepackage{verbatim}
\usepackage{graphicx}
\usepackage{balance}
\usepackage{array, makecell}
\setcellgapes{3pt}

\usepackage[hidelinks]{hyperref}
\usepackage{cmap}
\usepackage{overpic}
\usepackage{import}

\def\citep{\cite}

\usepackage{xcolor}

\title{Geometric Feature Prompting of Image Segmentation Models}
\author{Kenneth Ball, Erin Taylor, Nirav Patel, Andrew Bartels, Gary Koplik, James Polly, Jay Hineman
\thanks{K.~Ball (email: \url{kenneth.ball@geomdata.com}), E.~Taylor, G.~Koplik, and J.~Polly are with Geometric Data Analytics, Inc., Durham, NC, USA.
N.~Patel is with Penrose Research. A.~Bartels is with Georgia Institute of Technology. J.~Hineman is with Applied Research Associates.
}
\thanks{
This material is based upon work supported by the U.S.~Department of Energy, Office of Science, 
USDOE Office of Science (SC), Biological and Environmental Research 
under Award Number DE-SC-0020542.
}
\thanks{
The authors would like to acknowledge discussions and support from Dr.~Anjali Ayer-Pascuzzi of Purdue University.
}
\thanks{This work has been submitted to IEEE Transactions on Image Processing for possible publication.}
}

\begin{document}
\maketitle

\begin{abstract}
Advances in machine learning, especially the introduction of transformer architectures and vision transformers, have led to the development of highly capable computer vision foundation models.
The segment anything model (known colloquially as SAM and more recently SAM 2), is a highly capable foundation model for segmentation of natural images and has been further applied to medical and scientific image segmentation tasks.
SAM relies on prompts --- points or regions of interest in an image --- to generate associated segmentations.

In this manuscript we propose the use of a geometrically motivated prompt generator to produce prompt points that are colocated with particular features of interest.
Focused prompting enables the automatic generation of sensitive and specific segmentations in a scientific image analysis task using SAM with relatively few point prompts.
The image analysis task examined is the segmentation of plant roots in rhizotron or minirhizotron images, which has historically been a difficult task to automate.
Hand annotation of rhizotron images is laborious and often subjective; SAM, initialized with GeomPrompt local ridge prompts has the potential to dramatically improve rhizotron image processing.

The authors have concurrently released an open source software suite called geomprompt (\url{https://pypi.org/project/geomprompt/}) that can produce point prompts in a format that enables direct integration with the segment-anything package.

\end{abstract}


\section{Introduction}
Digital image segmentation --- the task of partitioning pixels into sets that differentiate image components or characteristics --- is a broadly-scoped computer vision task exhibiting both a diversity of 
techniques and a correspondingly diverse range of applications.

Some image segmentation tasks are simply defined and amenable to relatively non-ambiguous algorithmic processing. 
For instance, identifying the sets of pixels between level-set contours of gray-scale images is a task that can be accomplished through simple algorithmic processing. 
Likewise, algorithms utilizing geometric properties of images, such as ridge-like features, can be utilized to produce segments and masks of image components that exhibit ``tubular'' structures \citep{frangi1998multiscale}.
Watershed segmenters can likewise segment local basins in gray-scale images \citep{vincent1991watersheds, edelsbrunner2009persistent}.

Other segmentation tasks, for instance semantically specified tasks in natural images \citep{arbelaez2012semantic}, are well-specified in a qualitative sense but may not be sufficiently amenable to hand-derived algorithmic segmentation approaches.
Identifying and differentiating groups of pixels in natural images as ``cars'' or ``cats'' is a reasonable task for a human actor, but it is unlikely that a hand-derived algorithm will reliably accomplish these segmentation tasks in the diversity of configurations in which the target objects might appear.

For these problems, learned approaches provide an increasingly viable means of rapid segmention of semantically meaningful objects \citep{ronneberger2015u, garcia2017review}. 
We refer to such approaches as ``learned'' rather than AI to emphasize the role that training/optimization plays in their construction.
The advent of transformer models \citep{vaswani2017attention} has enabled the development (through large-scale training efforts) of foundation models that are trained on large, relatively heterogenous datasets and can subsequently be used to support wide varieties of downstream tasks \citep{bommasani2021opportunities}.
For digital image segmentation, the Segment Anything Model (SAM) \citep{kirillov2023segment} is especially notable for its wide applicability and compositionality by design.

In this effort we explore SAM's transference to segmentation problems that fall outside of the natural image paradigm of its primary training dataset (SA-1B) and that exhibit fine structures which SAM may overlook.
In particular, we are motivated by a prominent segmentation problem in the plant sciences: the differentiation of fine plant root structures from soil backgrounds and other artifacts within \emph{in situ} images.

Our approach leverages geometric features of plant roots to create focused prompts for SAM mask inference.
Observing that plant roots exhibit ridge-like structures, we produce a custom Python implementation of the multi-scale ridge detection algorithm \citep{lindeberg1998edge} and utilize local ridge features to generate sets of prompts for SAM segmentation.
We compare geometric prompting to uniform grid and random point prompts within SAM.
We also make a comparison between the collaborative geometric prompting and SAM segmentation and a custom algorithmic segmentation routine. 

Since the work discussed in this effort was completed an improved version of SAM, SAM 2 \cite{ravi2024sam} was released.
As the present manuscript introduces a novel use of differential geometry for prompting, the results presented here are easily translatable to and valid in a SAM 2 context and are expected to generalize to other point prompted segmenters. 

\section{Geometric Prompting of SAM}
\subsection{Algorithmic Prompting}
Segmentation routines that solely leverage quantifiable geometric features of images can be relied upon to behave reproducibly up to certain parameter specifications (thresholds, sensitivity parameters, etc.) and results are traceable to and explainable according to the specific segmentation algorithms.
Such approaches are useful when seeking to produce segmentations of image components that exhibit such quantifiable properties against backgrounds or extraneous artifacts that do not share these properties. 
For instance, the segmentation of blood vessels in medical images motivated the development and use of the Frangi vesselness filter referenced earlier \citep{frangi1998multiscale}.
Likewise, the topological watershed is an algorithm for segmenting grayscale images \citep{couprie1997topological, couprie2005algorithms, edelsbrunner2010computational} that has been applied to cell segmentation \citep{hu2021topology}.

We differentiate this ``algorithmic'' segmentation from ``learned'' segmentation: the latter involves the training of deep networks to functionally approximate segmentation decisions made by hand over large sets of image examples.
Consider the task of segmenting distinct cells in biological microsopy images.
Watershed segmentation is an algorithmic approach that leverages known geometric properties of microscopy images of cells to produce a solution to the segmentation problem.
The deep convolutional U-net architecture was developed to produce a learned solution to the same problem of cell segmentation \citep{ronneberger2015u}; while algorithmic choices are made in design and training, the actual segmentation task emerges as a learned functional reproduction of many labeled examples exemplifying ``ground-truth'' solutions.
Watershed and a trained U-net model are exemplars of algorithmic and learned solutions to the similar cell segmentation tasks.

Foundation models present new opportunities for collaboration between algorithmic and learned image processing.
Recognizing that image segmentation is a broad application area, SAM in particular has been designed with the goal of producing ``a broadly capable model that can adapt to many (though not all) existing and new segmentation tasks via prompt
engineering'' \citep{kirillov2023segment}.
SAM admits both sparse (points, bounding boxes, text) and dense (mask) prompts, but otherwise relies on a collaborative user or algorithm to provide meaningful prompts.
SAM's fully automatic segmentation mode is prompted by a uniform grid of points along with a hierarchy of rectangular cropped masks, followed by a mask filtering and disambiguation process.

\begin{figure}
\centering
\subfloat[Ridge prompts]{
	\label{ex1:a}
	\includegraphics[width=0.45\linewidth]{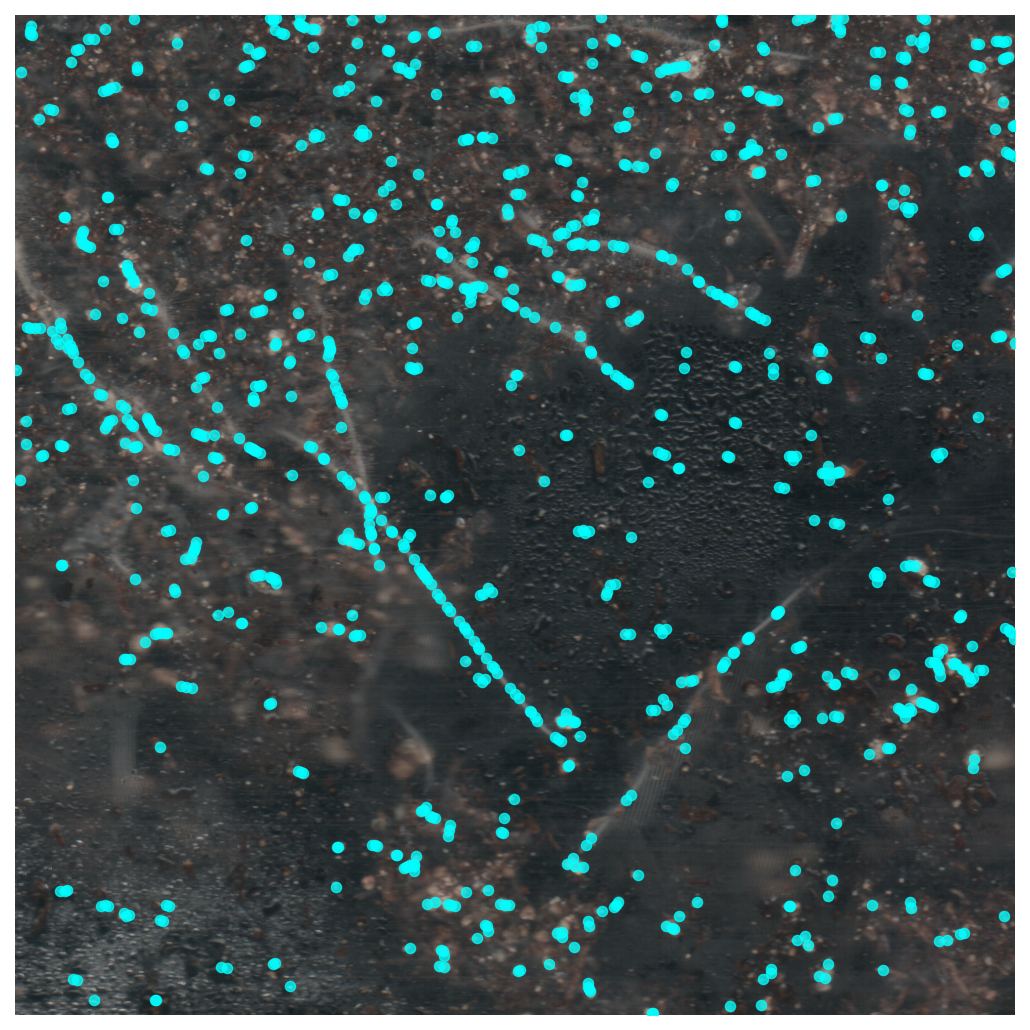}
}
\subfloat[Grid prompts]{
	\label{ex1:b}
	\includegraphics[width=0.45\linewidth]{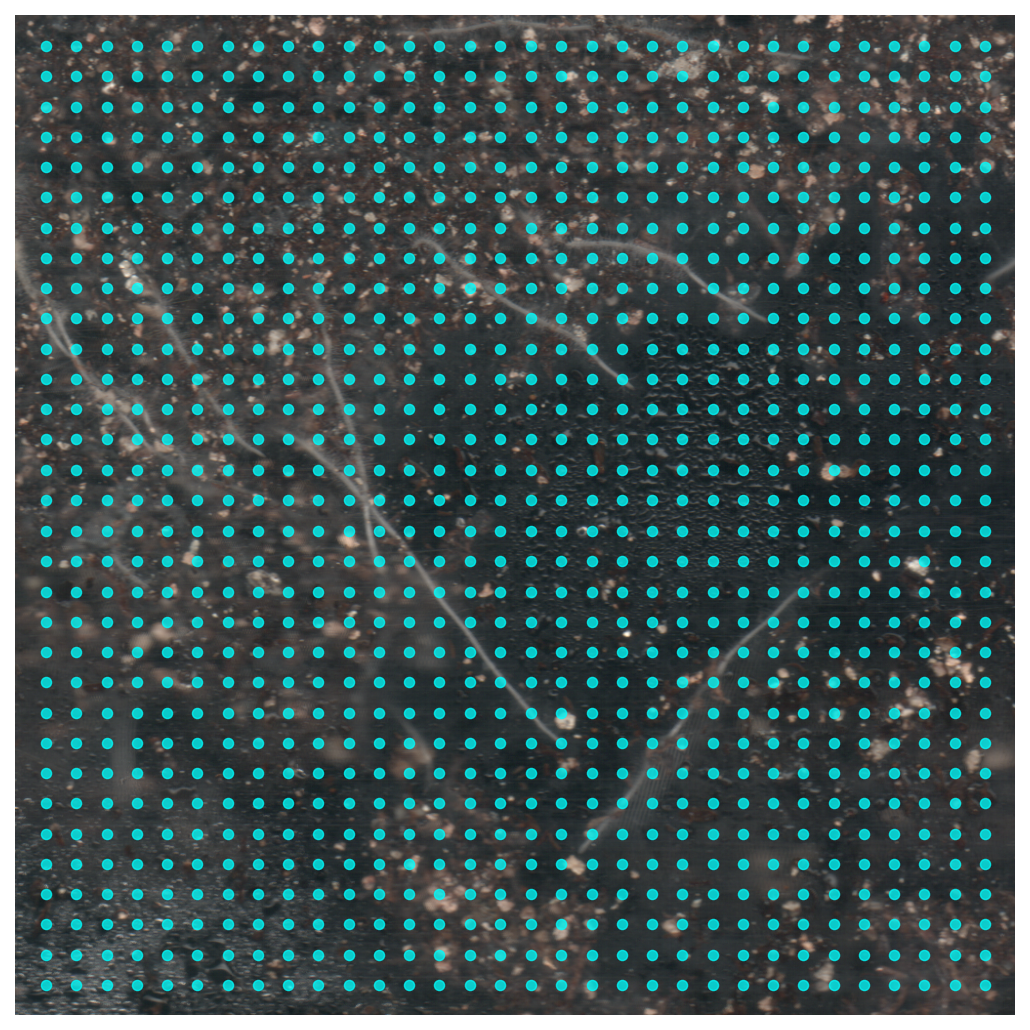}
}\\
\includegraphics[width=0.75\linewidth]{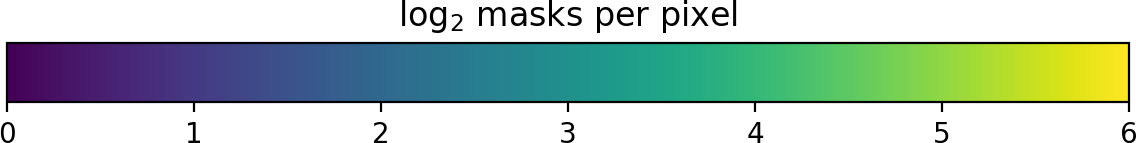} \\
\subfloat[Ridge SAM masks]{
	\label{ex1:c}
	\includegraphics[width=0.45\linewidth]{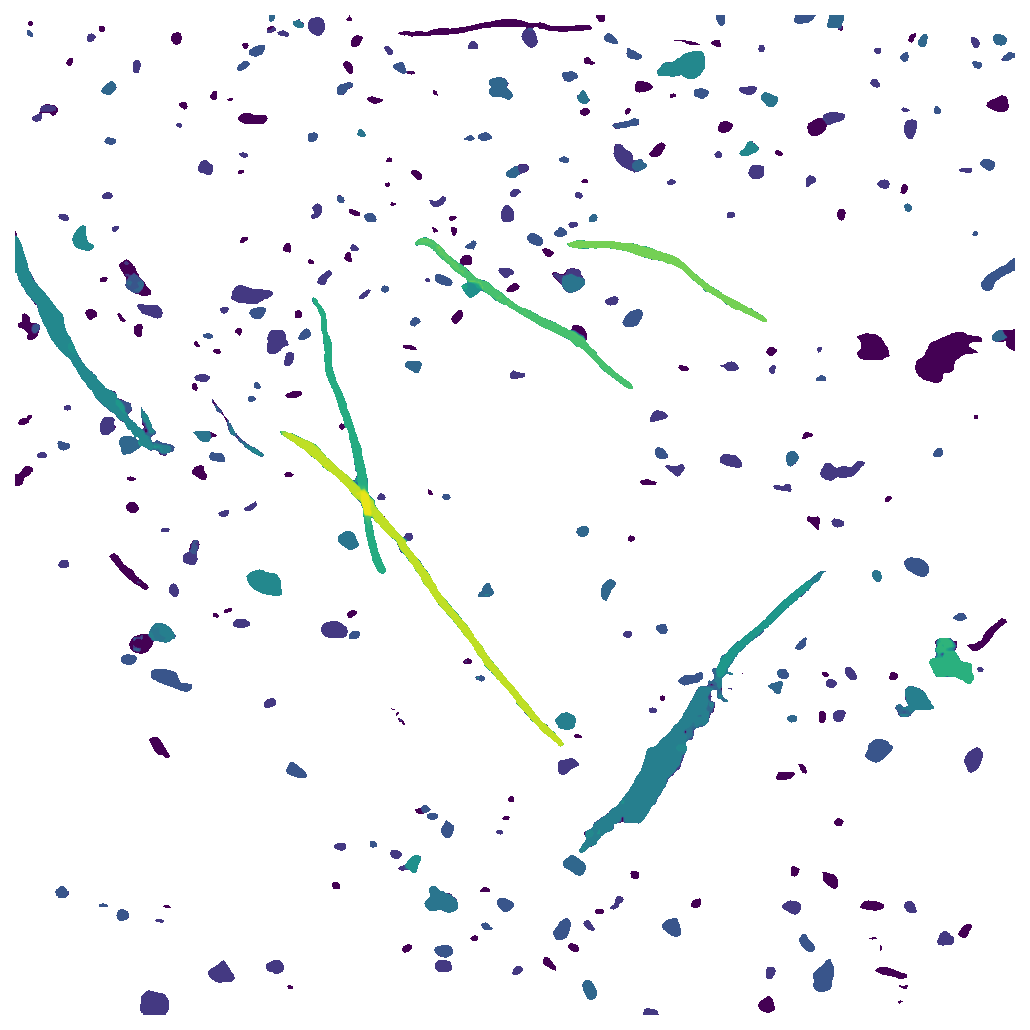}
}
\subfloat[Grid SAM Masks]{
	\label{ex1:d}
	\includegraphics[width=0.45\linewidth]{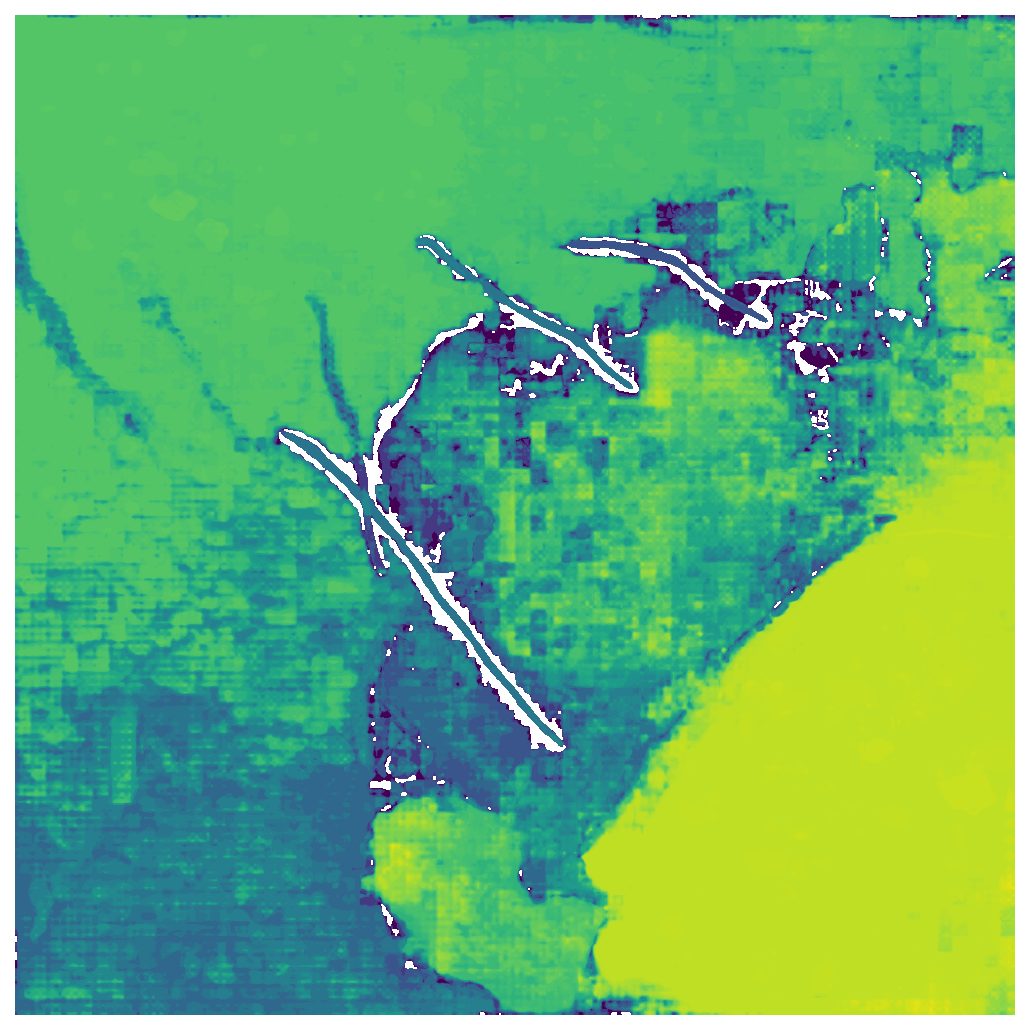}
}
\caption[]{A $1024\times 1024$ minirhizotron image sample with \subref{ex1:a} 1001 ridge point prompts and \subref{ex1:b} 1024 uniform grid point prompts. By prompting SAM in with local ridge like features, resulting good quality (predicted IoU $>0.75$) segmentation masks are much more focused on objects exhibiting those features \subref{ex1:c} than masks generated with naive grid points \subref{ex1:d}.}
\label{ex1.fig}
\end{figure}

Pandey et al.~\citep{pandey2023comprehensive} have investigated the integration of a learned prompter with SAM.
In this work we investigate and implement an \emph{algorithmic} prompting using local geometric features that enables SAM to be utilized for a specialized image segmentation task.
We illustrate the utility of this approach in Figure \ref{ex1.fig}: prompts that are focused on features of interest in a segmentation task --- in contrast to a comparably sized set of evenly distributed point prompts --- result in masks that are much more attentive to those features.
Our approach is comparable to a topological data analysis (TDA) derived point prompting recently suggested by Glatt and Liu \cite{glatt2023topological}: extrema filtered via persistent homology will provide contextually useful point prompts to a segmenter seeking (spatial) scale invariant coherent intensity features.
Like Glatt and Liu we use geometric features to generate non-learned prompts, and we affirm that there are many opportunities of for using non-learned geometric and topological image structures for enhanced segmentation.
We point out that these non-learned prompting strategies are critical especially for applications --- like rhizotron image processing --- where factors like constrained investment in labeling, increased heterogeneity in images, and complexity in hand annotation conspire to limit quantity and quality of labeled data relative to need.

The ridge prompting presented here is more tailored to the context of root segmentation, as evidenced in Figure \ref{ex1:c} by greater ``attention'' paid to elongated root features relative to more circular luminous soil artifacts.
In the remainder of this section we describe the rhizotron image processing problem in greater detail and we describe the derivation of ridge-like features.

\subsection{Rhizotron Image Processing}
In particular, we are motivated by a prominent segmentation problem in the plant sciences: the differentiation of fine root structures from soil backgrounds and other artifacts in rhizotron images.
Rhizotrons are transparent interfaces with soil systems via which images can be taken of plant roots \emph{non-destructively} \citep{huck1982rhizotron, mcmichael1987applications}; somewhat like an ant-farm but for plant roots.
Rhizotrons come in a variety of form factors and minirhizotrons --- transparent tubes that can be inserted into the ground or in series of containers --- constitute a form factor that is widely used for \emph{in situ} experimentation \citep{iversen2012advancing, vamerali2012minirhizotrons, childs2019spruce}.

Automation of the processing of such images, especially segmentation of roots and their subsequent registration and tracking has been a subject of an intense interest, and prompted the research under which the present work has been funded.
The root segmentation task is a significant bottleneck in the conduction of rhizotron-style experiments.

This task has generally been accomplished by hand-tracing, even as image collection systems have progressed over the better part of a century \citep{bates1937device, waddington1971observation}; Cheng et al.~\citep{cheng1991measuring} even describe tracing of VCR recorded stills displayed on a CRT TV screen.
Even with the advent of widely available digital imaging and software enabling direct interaction with images, segmentation has still generally been accomplished by lab technicians tracing individual via a user interface---frequently with some algorithmic assistance to infer root widths, e.g.~RootFly \citep{zeng2008automatic} or WinRHIZO Tron as used in \citep{shen2020high}.

More recently, especially with the advent of deep convolutional neural network (CNN) architectures like U-net, the training of a variety of models for the segmentation of rhizotron-style images has been reported \citep{wang2019segroot, smith2020segmentation, shen2020high, narisetti2021fully, peters2023good, baykalov2023semantic}.
However, labeled data (hand annotated/segmented) is relatively expensive to obtain, especially in relation to the heterogeneity of soil substrates, root systems, experimental apparatus, lighting conditions, and artifacts that might appear in diverse rhizotron-style experiments.
Rhizotron segmentation model training have been observed to benefit from transference of knowledge: Xu et al.~\citep{xu2020overcoming} report root segmentation accuracy improvements using a model pretrained on a cross-species segmentation task, but fine-tuning in this case still involves training on a significant proportion of hand-annotated segments in the target task.

Baykalov et al.~\citep{baykalov2023semantic} provide an investigation into the comparative use of backboned U-net segmenters and other learned segmenters with and without in sample fine-tuning via data augmentation. approaches. 
We note that the authors report meaningfully reduced model performance (illustrated by AUC-ROC curves) when base and augmented models are applied to an unseen out-of-species experiment.

The heterogeneity of soil backgrounds, diversity of experimental settings and artifacts, the wide range of plant species studied, and the relative expense of obtaining hand-annotated data all point towards challenges in training a truly automated and generalizable learned segmentation solution for rhizotron-style images.
We observe that U-net architecture semantic segmenters with existing backbones can generally achieve ``good'' (IoU) performance when augmented with in-experiment annotation data, however this still necessitates expensive hand-annotation for each new experimental design even if does reduce the scope of work from the entire set to some subset of images.

We therefore arrive at our motivation for adapting SAM to rhizotron image processing with geometric prompting: foundation models exhibit the ability to rapidly adapt to a new task by adding model components and data that are fractional to the pretraining of the foundation model itself.
This geometrically motivated augmentation of SAM adapts the foundation model to enable efficient sensitivity to specific fine scale structures of interest \emph{and} outputs segmentations that tend to be more accurate than benchmark hand annotations. 
Shaharabany et al.~\cite{shaharabany_autosam_2023} demonstrate that SAM is an effective foundation model; they replace SAM's image encoder with a custom encoder that produces image prompts for composition with SAM (the \emph{Auto} adjective is a reference to automating the prompting of SAM).

Moreover, geometrically prompted SAM yields a feasible \emph{instance} segmenter for roots (CNN segmeters like trained U-net are generally semantic segmenters). 
Instance segmentation is highly relevant for nondestructive root image processing, wherein growth and turnover of root features is of significant interest. 

\subsection{Multiscale Ridge Detection}\label{murid.sec}

Consider a grayscale digital image of pixel dimensions $M \times N$. A ridge detector seeks to segment the image by differentiating ``ridge-like'' (or ``valley-like,'' depending on the relative luminosity of targets) regions from a general heterogeneous background. 
In the language of our motivating root-soil example, we would like to segment distinct roots from soil background by taking advantage of the observation that, in rhizotron images, roots tend to be both \emph{bright and elongated} relative to soil backgrounds.

We utilize the methodology presented by Lindeberg \citep{lindeberg1998edge} to find locally ridge-like pixels. While we point the reader to their paper for a full exposition, here we present a few details meant to illustrate our implementation of the multi-scale ridge detection algorithm.

In an idealized setting, a two-dimensional image is a twice-differentiable function $f:\mathbb{R}^2 \to \mathbb{R}$. Its scale space representation is the convolution of the image with a scale-varying Gaussian kernel:
\begin{equation}
L(x, y; t) = \left[\frac{1}{2\pi t}e^{-(x^2 + y^2)/(2t)}\right] \star f(x, y).
\end{equation}
There are two important things to note. First, the scale space representation emphasizes larger scale image features as $t$ grows because small scale features are blurred out. Second, this definition of a scale space representation is differentiable in the scale direction $t$, allowing for testing for critical points in scale space.

A variety of tests of local ridge strength can be designed by leveraging the observation that principal curvatures (eigenvalues of the Hessian of $f$) are descriptive of local ridge-like structures: the direction along the ridge should have principal curvature close to zero, whereas the direction transverse to the ridge should exhibit a significantly negative principal curvature.
We utilize the \emph{square of the $\gamma$-normalized principal curvature difference} of the scale space image $L$ defined \citep{lindeberg1998edge} as 
\begin{equation}\label{pcd.eqn}
\mathcal{A}L = t^{2\gamma}\left(
	(L_{xx}- L_{yy})^2 + 4 L^2_{xy}
\right)
\end{equation}
as an indicator of ridge strength that is less sensitive to blobs (e.g.~non-elongated luminous structures). $\gamma$ is a normalizing parameter that can be set to $\gamma=\tfrac{3}{4}$ for compatibility with the width of a cylindrical ridge. For comparison purposes we also include an option to utilize the ``square of the $\gamma$-normalized \emph{square} principal curvature difference'' as the ridge test value in our accompanying code release; see Lindeberg \citep{lindeberg1998edge} for details.

We can use the ridge intensity test to determine points in the domain of the scale-space image that are ridge-lines. This is the set of points $(x, y; t)$ where, in addition to being locally ridge-like according to their principal curvatures, there is a zero-crossing of the gradient in the direction of most negative curvature and where the ridge intensity test achieves a local maximum in scale space. 
Of course, digital images are discrete structures, so in practice these tests are implemented using discrete derivatives.

Given a digital image of dimensions $(M, N)$ and a series of $K$ scales, the scale-space ridge test results in a sparse array $R$ of dimension $(M, N, K)$ where non-zero pixels (those that satisfy the ridge criteria above) are assigned their local ridge test value.
This skeleton-like structure approximates curves sweeping through the domain of $L$ representing ridge-lines along the ``top'' of ridge features of varying widths.

Because it is dependent on principal curvature values, this ridge test value breaks down at local saddle surfaces of third-order or higher (e.g. the monkey saddle) which can correspond to branchings and crossings of roots in actual images.
However, a generalized segmenter like SAM can still adequately segment these regions from connected root features that exhibit local differential geometry amenable to the ridge intensity test.

We utilize a subset of these ridge-like points --- selected via a filtering algorithm described below --- as point prompts for SAM.
We then assess the performance of automated segmentation utilizing these geometric point prompts on a benchmark minirhizotron dataset.




\section{Methodology and Experiments}
\subsection{Point Prompt Selection}\label{point_prompt_selection.sec}

We seek to further filter the sparse array of scale-space ridge points $R$ into a set of $K$ discrete point prompts in the two-dimensional image coordinate space. 
There are many ways that this filtering could be accomplished: a simple approach is to sort scale-space ridge points by their ridge test value $\mathcal{A}L$ \eqref{pcd.eqn} and then take the image space coordinates of the $K$ largest values as the set of point prompts.
However, this test value filtering approach would tend to cluster selected points in image regions of high intensity, creating unreliable redundancy in prompting for certain very ridge-like features while potentially ignoring meaningful, if somewhat fainter, features.
Alternatively, the non-zero scale-space ridge points could simply be randomly sampled $K$ times: while this approach would serve to distribute prompts throughout the image it may tend to significantly overemphasize artifactual ridge-like features.

These dichotomous approaches --- random sampling and sorting purely by point-wise ridge test values --- neglect to assess locally aggregate properties of pixel-connected features.
Instead we turn again to Lindeberg \cite{lindeberg1998edge} for a definition of ridge \emph{salience}, which is the path integral of the root of the test value of a connected scale-space ridge curve $(x, y, t) \in \Gamma$ projected onto the image space:
\begin{equation}\label{salience.eqn}
A(\Gamma) = \int_{(x, y) \in \text{proj}\Gamma} \sqrt{\mathcal{A}L(x, y, t)} ds.
\end{equation}
We identify connected scale-space ridge curves in the sparse array $R$ and compute a discrete approximation of ridge salience for each component curve.

While we could at this point return a representative point prompt for each ridge curve, sorted by salience, we instead seek to further balance such a distribution of prompts with the notion that more ``attention'' should be paid to more salient features.
Thus, we allocate $K$ point prompts to ridge curves \emph{in proportion} to their relative salience: given $n$ ridge curves sorted by salience such that $A(\Gamma_1) \leq A(\Gamma_2) \leq \ldots \leq A(\Gamma_n)$ then each curve is assigned $\lceil K\cdot A(\Gamma_i) / \sum_j^n A(\Gamma_j) \rceil$ ridge points, which are selected randomly from each within each curve in descending salience order until the specified number of prompts $K$ are generated.
We recognize that alternate algorithms for filtering and selecting point prompts based on ridge salience could be also be explored and utilized.
\subsection{Benchmark Dataset} 
We used a subset of minirhizotron images from the plant root minirhizotron imagery (PRMI) dataset~\cite{xu2022prmi}. 
This dataset was primarily created for plant root segmentation tasks and contains root images from several plants. Every image has an image-level annotation that indicates whether the image contains a root, and some images have pixel-level annotations (i.e., binary masks) that indicate whether each pixel is a root.

\begin{figure}
\centering
\includegraphics[width=1.0\linewidth]{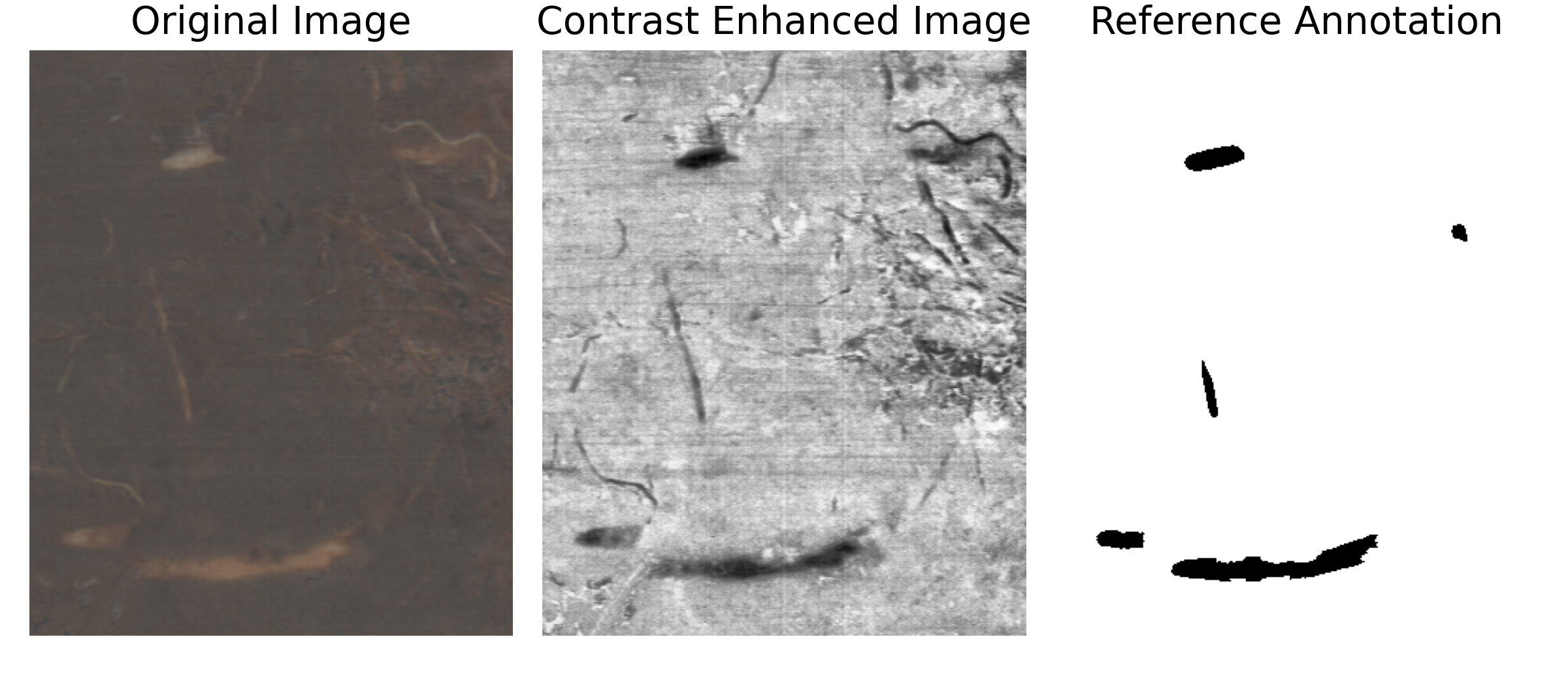}
\caption{An example of benchmark minirhizotron image (left), a contrast enhanced version (center), and an the accompanying annotation of root segments. Significant ridge-like features of varying scales are not included in the reference annotation, which we note not to impugn annotation quality in this particular case but rather to illustrate that the geomprompt + SAM objective (segment ridge-like features) differs from the benchmark analysis objective (segment plant roots of interest). Also, contrast enhancement is used here to illustrate faint fine features of potential interest, but was not used in our experimental analysis.}\label{ex_sg.fig}
\end{figure}

In our experiments, we considered images of the switchgrass species that contain a root and are accompanied by binary masks. 
We chose the switchgrass images specifically because they have a characteristically fine, narrow root structure in this image collection paradigm. 
We observe that the annotation process utilized in the generation of this dataset yields visibly imperfect segmentations: pixel level annotations of root features are frequently inaccurate, and we likewise often notice features that appear to be roots but are not annotated (see Figure \ref{ex_sg.fig}).
We hesitate to speculate on the quality of the benchmark dataset we selected beyond these observations, which we note only to contextualize subsequent experimental results.
However, we assert that (even computer assisted) annotation of high resolution minirhizotron and rhizotron images is a notoriously tedious and subjective task. 
As part of a broader research effort to automate rhizotron image processing we have yet to observe an objectively, consistently reliable hand annotated dataset at the experimental scale necessary to capture inherent variability in these image modalities.

In total, we generated segmentations for 2,419 switchgrass root MR images in the referenced experimental training set.

\subsection{Experiment Design}
In our experiment we sought to compare the results of geomprompt derived point prompts to uniform grid prompt points.
To produce uniform gridded prompts we used tools included in the segment-anything package to derive 16, 64, 256, and 1024 prompt points from $4\times 4$, $8\times 8$, $16 \times 16$ and $32 \times 32$ uniform grids overlaid on the image, respectively.  

As discussed in Section~\ref{point_prompt_selection.sec}, we likewise tuned the ridge detection algorithm to obtain an approximately correct number of geomprompt points at each scale for each image.
To initialize the SAM model for segmentation, we used the default model checkpoint.

In SAM, masks can be filtered according to two thresholds: \texttt{pred\_iou\_thresh} and \texttt{stability\_score\_thresh}. Both thresholds are real values in $[0, 1]$. 
The former depends on the model's predicted mask quality, while the latter corresponds to the stability of the mask under changes to the cutoff that binarizes the model's mask predictions.
If the score of a mask is below at least one of the user-specified values for these thresholds, the mask is filtered out. 
In our experiments, we set \texttt{pred\_iou\_thresh} to $0.6$ and \texttt{stability\_score\_thresh} to $0.8$. 
These values are lower than their default values of $0.88$ and $0.95$, respectively, because we are willing to tolerate masks that are non-root artifacts in exchange for capturing all roots in an image.

We additionally filtered segments by area (in proportion to the total image area), noting that point prompts in soil backgrounds may result in segment regions that encompass broad areas of the image (see the large background segment regions resulting from grid prompt points in Figure \ref{ex1:d}).
Roots are fine structures, and affirming that no reference annotation in the training set had --- in aggregate --- more than 25\% of pixels labeled as switchgrass roots, we set a threshold of 25\% of the image area above which we would filter segments from further consideration.

\subsection{Results}

\subsubsection{Segmentation Quality}
\begin{figure}
	\centering
	\includegraphics[width=1.0\linewidth]{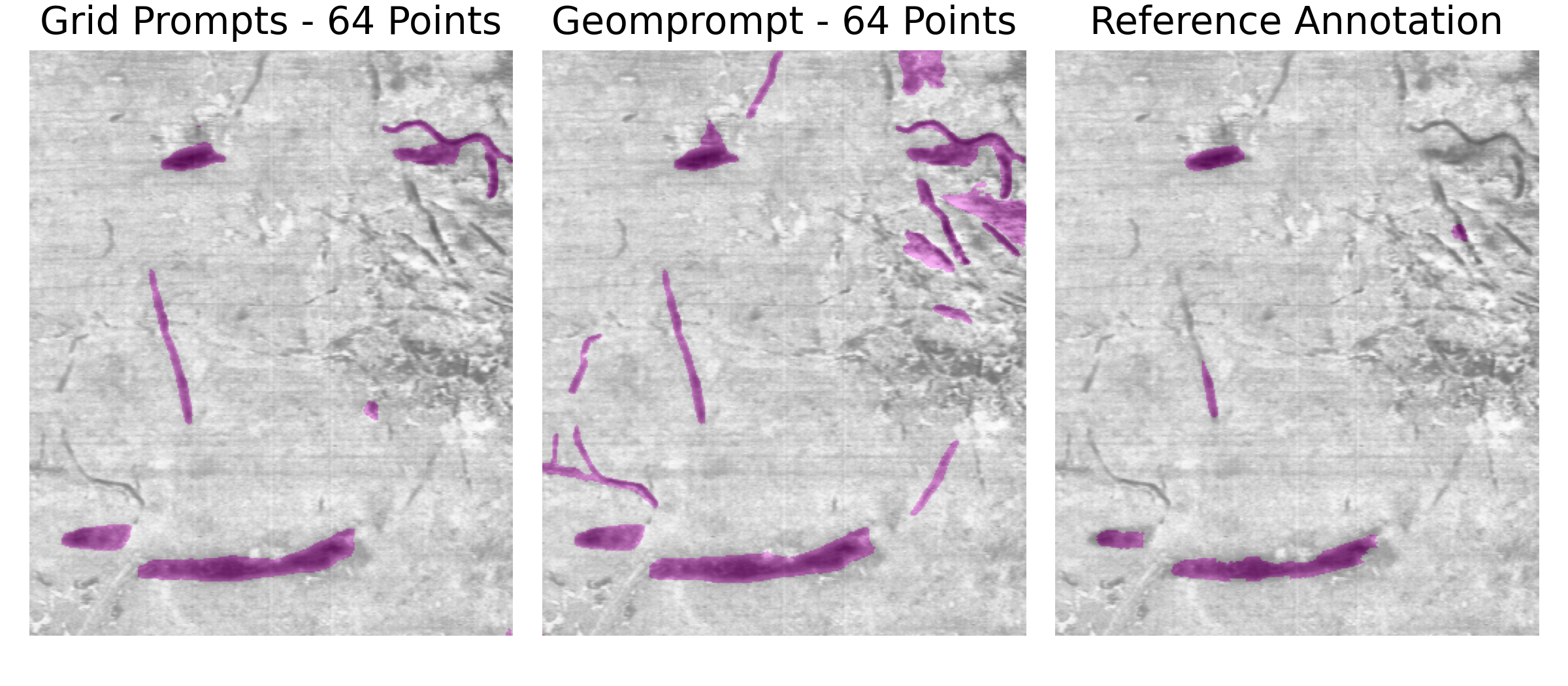}
	\caption{Examples of 64 point density grid prompted SAM segmentation (left), geomprompt SAM segmentation, and the benchmark reference annotation, overlaid on a contrast enhanced image detail.}\label{segs.fig}
\end{figure}

As discussed immediately above, segments were filtered according to SAM predicted IoU, SAM stability, and absolute segment area relative to the overall image area.
At lower prompt densities, geomprompt points tended to produce better ``quality'' SAM segments under the above threshold criteria than uniform grid prompts: at 16 prompt points 74\% of geomprompt points resulted in segments that met the threshold criteria as opposed to 47\% of grid prompt points.
At high density (1024) prompt points geomprompt produced slightly fewer quality segments than uniform grid prompting (about 2\% less); our point selection algorithm tries and distributes prompt points to less salient ridge features once more salient features are accounted for, inducing more transient feature prompting as the prompt density exceeds meaningful features for segmentation.
See Table \ref{prompt_quality.tab} for specific percentages of SAM segments that pass the thresholds.
A representative example of segmentation results is plotted in Figure \ref{segs.fig}.

\begin{table}
	\centering
	\makegapedcells
\begin{tabular}{r | c | c | c | c}
	~ & \multicolumn{4}{c}{Prompt Point Density} \\
	Prompt Method & 16 & 64 & 256 & 1024 \\
	\hline 
	Geomprompt & 73.9\% & 54.3\% & 34.2\% & 16.6\% \\
	Grid prompt & 46.8\%  & 43.7\% & 31.8\% & 18.4\% 
\end{tabular}
\caption{Percentage of prompted segments that pass predicted IoU, stability, and maximum area thresholds.}\label{prompt_quality.tab}
\end{table}

\subsubsection{Segmentation Analysis Relative to Annotated Masks}
We next compared quality filtered SAM segmentations against the benchmark annotation masks.
We report true positive rates (TPR's) and false positive rates (FPR's) at the pixel level, aggregated across all images in the training set in Table \ref{prompt_rates.tab}.
TPR's indicate that geomprompt points can capture the majority (nearly 70\%) of benchmark labeled pixels with an accompanying low FPR of around 5\%.
In this experimental dataset geomprompt + SAM TPR increases marginally beyond a density of about 64 prompt points, while FPR's continue to rise more rapidly at higher densities.

Uniform grid prompting with SAM, on the other hand, requires about an order of magnitude more prompt points to achieve TPR's comparable to geomprompt while consistently returning higher FPR's.
This is as expected: geomprompt is leveraging local differential geometry to orient the SAM segmentation prompts towards features that are more likely to be roots, whereas deriving prompts from a uniform grid effectively relies on random chance for an annotated root feature to coincide with a prompt point (we posit that random sampling of prompt points from a 2-D uniform distribution over the image space would result in comparable segmentation performance).

In addition to TPR and FPR's we report intersection over union (IoU) in this experiment, however we note that IoU is not particularly informative because of a large imbalance between benchmark negative (background) pixels and positive (root) pixels. 
Annotated root features are relatively sparse, generally encompassing less than 2\% of any given benchmark mask.
This is further exacerbated by the previously described objective mismatch and/or quality disparity between our segmentation task and that of the benchmark annotation.

\begin{table}
	\centering
	\makegapedcells
	\begin{tabular}{r | c | c | c | c}
		~ & \multicolumn{4}{c}{Prompt Point Density} \\
		Prompt Method & 16 & 64 & 256 & 1024 \\
		\hline \hline
		~ & \multicolumn{4}{c}{True Positive Rate (TPR)} \\
		\hline 
		Geomprompt & \textbf{68.1\%} & \textbf{77.6\%} & \textbf{80.9\%} & 82.7\% \\
		Grid prompt & 23.9\%  & 51.9\% & 75.2\% & \textbf{83.0\%} \\
		~ & \multicolumn{4}{c}{False Positive Rate (FPR)} \\
		\hline 
		Geomprompt & \textbf{6.4\%} & \textbf{16.7\%} & \textbf{28.9\%} & \textbf{37.4\%} \\
		Grid prompt & 12.4\%  & 22.1\% & 33.0\% & 41.2\% \\
		~ & \multicolumn{4}{c}{Intersection over Union (IoU)} \\
		\hline 
		Geomprompt & \textbf{0.180} & \textbf{0.088} & \textbf{0.047} & \textbf{0.035} \\
		Grid prompt & 0.031  & 0.040 & 0.036 & 0.031 
	\end{tabular}
	\caption{Pixel-level true and false positive rates of geompprompt and grid prompted SAM segmentations relative to the benchmark mask, aggregated across all experimental images.}\label{prompt_rates.tab}
\end{table}


\section{Discussion}
We have demonstrated that geomprompt, coupled with SAM, can produce segmentations more efficiently (with fewer prompt points) that better match a benchmark segmentation task (as measured by pixel-wise TPR and FPR) relative to a naively prompted SAM.
Efficiency is of interest because, while SAM produces individual segments quite quickly in response to prompts (following image encoding), disambiguation of overlapping prompts in a ``segment-everything'' mode may still be a computationally intensive task depending on how it is accomplished.
We caution that there is computational overhead in actual geomprompt computation of ridge-like prompts, which may negate any downstream efficiency gains in SAM and post-SAM processing depending on geomprompt parameters like scale range and resolution.

Much more importantly, we assert that we can improve the utility of SAM (and similar foundational segmentation models) as a tool in broader image analysis tasks by focusing its attention on relevant image features.
Envision a situation where a point-prompted SAM model is utilized to provide candidate segments for user-annotation: a segmentation methodology that can more reliably focus a user's attention to features that are likely targets of their segmentation task will yield much more important efficiency gains in the application of human and expert attention.
This is a motivating principle behind our development of the method reported here, which is part of a broader sem-automated minirhizotron image analysis effort wherein a subset of segments are meant to be passed for user feedback in an active learning cycle \citep{settles2009active}.

We further emphasize that focused prompting through geomprompt improves SAM's efficacy as a foundation model. 
In this case we have utilized SAM entirely ``out of the box,'' to effectively segment features in a paradigm that is in many ways very different from the natural image segmentation tasks it has been trained to accomplish.
Further segmentation performance gains could be realized by fine tuning released, check-pointed SAM models with either user labeled segments or even from the raw geomprompt segments themselves coupled with some additional filtering.
The approach described here as ``geomprompt'' also focuses primarily on ridge-like features, but the method can easily be extended to concurrently include inverted, valley-like features, or even other geometric features. For example, Lindeberg \cite{lindeberg1998edge} describes a very similar differential geometry procedure for identifying multi-scale edges in images, and geometric featurizations highlighting blobs or other structures could be designed depending on a desired segmentation task. 

\bibliographystyle{IEEEtran}
\bibliography{murid}

\end{document}